\documentclass[10pt,twocolumn,letterpaper]{article}

\usepackage[pagenumbers]{cvpr} 
\usepackage{multirow}
\usepackage{soul}
\usepackage{textcomp,booktabs}
\usepackage{amssymb}
\usepackage{pifont}
\usepackage{makecell}

\usepackage{booktabs}
\usepackage{amsmath}
\usepackage{amsfonts}
\usepackage{algorithmic}
\usepackage{dblfloatfix}
\usepackage{graphicx}
\usepackage{textcomp}
\usepackage{caption}
\usepackage{subcaption}
\usepackage{atbegshi}
\usepackage{xstring}
\usepackage{flushend}
\usepackage[normalem]{ulem}
\usepackage{tikz}
\usepackage{multirow}
\usepackage{color}
\definecolor{forestgreen}{RGB}{0,139,69}
\usepackage{xcolor}
\definecolor{citecolor}{HTML}{0071bc}
\definecolor{SeaGreen4}{RGB}{0,205,102} 
\definecolor{SlateBlue}{RGB}{106,90,205} 
\definecolor{DarkRed}{RGB}{178,34,34} 
\usepackage[colorlinks, linkcolor=red,  anchorcolor=DarkRed, citecolor=SeaGreen4]{hyperref} 
\usepackage{makecell}
\usepackage{mathrsfs}
\usepackage{soul}
\usepackage{colortbl}
\usepackage{pgfplotstable} 
\usepackage{pgfplots}
\pgfplotsset{compat=newest} 
\usepackage{pgffor} 
\usepackage{filecontents}
\usepackage{enumitem}
\usepackage{pifont} 

\definecolor{cvprblue}{rgb}{0.21,0.49,0.74}
\def\paperID{*****} 
\def\confName{CVPR}
\def\confYear{2025}

\title{ R2GenKG: Hierarchical Multi-modal Knowledge Graph for LLM-based Radiology Report Generation }

\author{Futian Wang$^{1}$, Yuhan Qiao$^{1}$, Xiao Wang$^{1}$\thanks{Corresponding Author: Xiao Wang (xiaowang@ahu.edu.cn)}, 
    Fuling Wang$^{1}$, Yuxiang Zhang$^{1}$, Dengdi Sun$^{1}$ \\ 
${^1}${School of Computer Science and Technology, Anhui University, Hefei, China} \\ 
\small{\textit{\{e24301191, e23201049\}@stu.ahu.edu.cn}, \textit{\{wft, xiaowang\}@ahu.edu.cn}, \textit{z1169647007@foxmail.com}, \textit{sundengdi@163.com}}   
}

\begin{document}
\maketitle

\begin{abstract}
X-ray medical report generation is one of the important applications of artificial intelligence in healthcare. With the support of large foundation models, the quality of medical report generation has significantly improved. However, challenges such as hallucination and weak disease diagnostic capability still persist. In this paper, we first construct a large-scale multi-modal medical knowledge graph (termed M3KG) based on the ground truth medical report using the GPT-4o. It contains 2477 entities, 3 kinds of relations, 37424 triples, and 6943 disease-aware vision tokens for the CheXpert Plus dataset. Then, we sample it to obtain multi-granularity semantic graphs and use an R-GCN encoder for feature extraction. For the input X-ray image, we adopt the Swin-Transformer to extract the vision features and interact with the knowledge using cross-attention. The vision tokens are fed into a Q-former and retrieved the disease-aware vision tokens using another cross-attention. Finally, we adopt the large language model to map the semantic knowledge graph, input X-ray image, and disease-aware vision tokens into language descriptions. Extensive experiments on multiple datasets fully validated the effectiveness of our proposed knowledge graph and X-ray report generation framework. The source code of this paper will be released on \url{https://github.com/Event-AHU/Medical_Image_Analysis} 
\end{abstract}

\section{Introduction} 

In recent years, automated X-ray medical report generation~\cite{messina2022survey} has attracted increasing attention due to its potential to significantly enhance the efficiency and accuracy of radiological diagnosis compared to traditional manual reporting approaches. This task typically involves developing machine learning models that can perceive and interpret radiographic images through a vision encoder, and subsequently generate coherent, clinically relevant descriptions of pathological findings or diagnostic impressions using a language decoder. By integrating visual understanding with natural language generation, such systems aim to assist radiologists in producing consistent and comprehensive reports, thereby reducing workload and minimizing diagnostic oversights.

\begin{figure}
\centering
\includegraphics[width=0.48\textwidth]{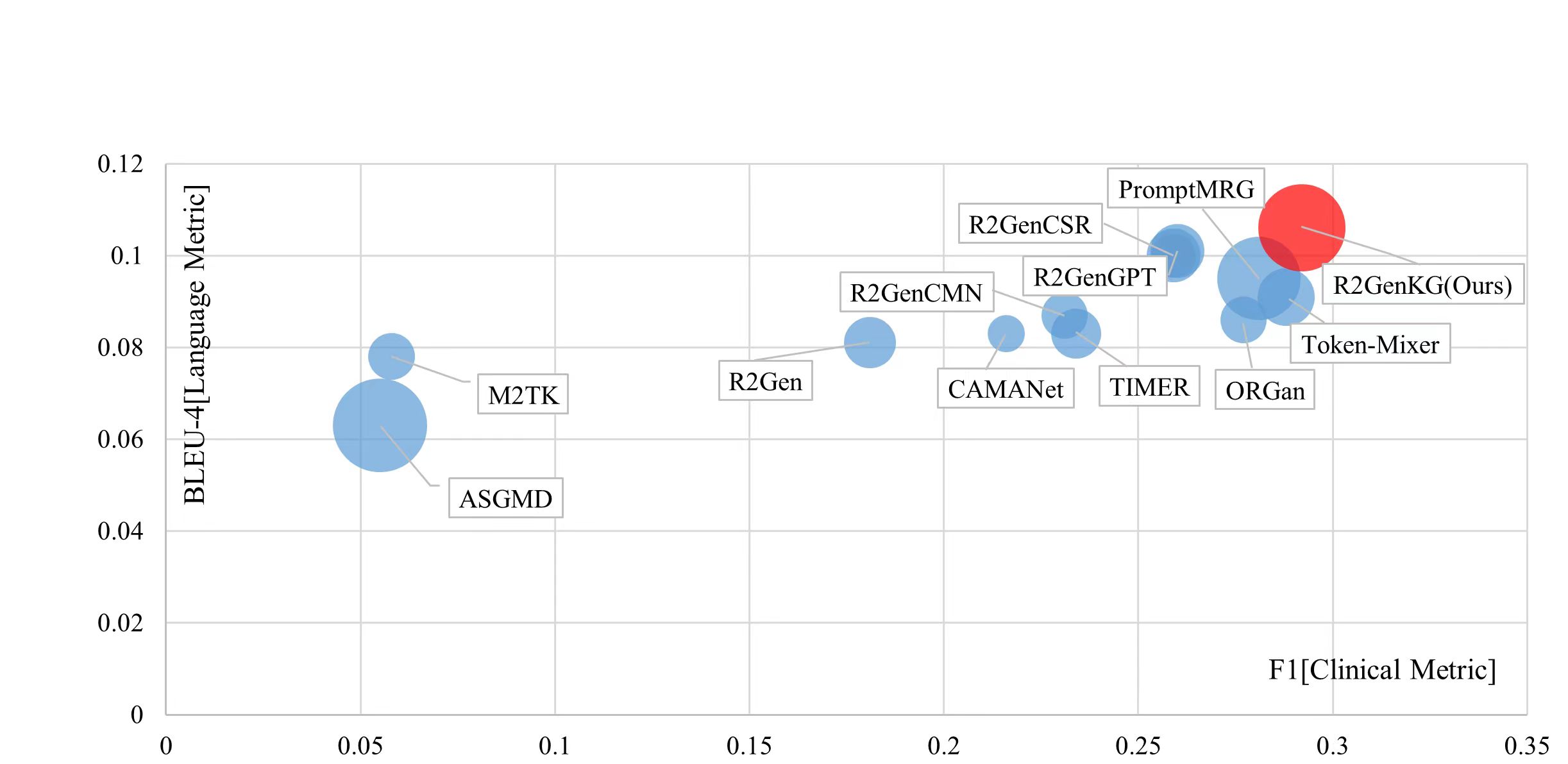}
\caption{An overview of the existing benchmark models on the CheXpert Plus dataset, the size of the bubbles represents the number of parameters.}
\label{fig:KG_Bubbles}
\end{figure}

With the development of Large Language Models (LLMs)~\cite{touvron2023llama2, yang2025qwen3, guo2025deepseekR1}, the generated report also achieves improved performance, such as R2Gen-GPT~\cite{wang2023r2gengpt}, CoFE~\cite{li2024CoFE}, MambaXray-VL~\cite{Wang_2025_CVPR}, R2GenCSR~\cite{wang2024r2gencsr}, and AM-MRG~\cite{wang2025AMMRG}. 
Specifically, Liu et al.~\cite{liu2024boostrapping} propose bootstrapping LLMs for RRG with an in-domain instance induction and a coarse-to-fine decoding manner. 
Li et al. propose the CoFE~\cite{li2024CoFE}, which learn non-spurious visual representations by contrasting the representations between factual and counterfactual images. 
Wang et al. propose a three-stage pre-trained Mamba network MambaXray-VL~\cite{Wang_2025_CVPR} for the X-ray medical report generation. 
To address hallucinations and insufficient disease diagnostic capabilities of large language models for report generation, some studies have attempted to incorporate knowledge graphs to enhance the quality of report generation~\cite{li2023DCL, liu2021auto}. 
Specifically, a dynamic knowledge graph is constructed and integrated with contrastive learning to enhance the model's understanding of chest X-ray images in DCL~\cite{li2023DCL}. KGAE~\cite{liu2021auto} leverages a pre-constructed medical knowledge graph to map images and texts into a shared semantic space, and employs a graph-enhanced decoder to generate more accurate medical reports.

\begin{table*}
\centering
\caption{Comparison between existing medical knowledge graph and our newly proposed M3KG.} 
\label{tab:KG_list}
\resizebox{\linewidth}{!}{
\begin{tabular}{l|l|c|c|c|c|c} 
\hline \toprule [0.5 pt]
\textbf{Title} &\textbf{Publish}  &\textbf{\#Entity}  &\textbf{Entity Classes}  &\textbf{Relations}  &\textbf{Construction}  &\textbf{Multi-modal} \\
\hline 
\#01 RadGraph~\cite{Jain2021RadGraphEC}  &NeurIPS 2021  &14,579  &Anatomical and Observational entity  &3 (suggestive of, modify, located at)  &Manual \& Machine  &\ding{55} \\
\hline 
\#02 SentSAT+KG ~\cite{zhang2020radiology} &AAAI 2020  &27  &Disease \& Organ categories    &\makecell[c]{Pairwise connected based on \\ co-occurrence relationships}  &Manual \& Machine  &\ding{55} \\
\hline 
\#03 DCL~\cite{li2023DCL} &CVPR 2023  & 28 + dynamic nodes &\makecell[c]{Disease keywords, organ names, \\ and root node}  & \makecell[c]{Co-occurrence +3 \\ (suggestive of, modify, located at)}  & Machine &\ding{55}   \\
\hline 
\#04 DCG~\cite{liang2024divide} &ACM-MM 2024  & IU-Xray: 191 ; MIMIC-CXR:276   & Normal/Abnormal entity pairs. & exists , not exist & Machine  &\ding{55}  \\ 
\hline 
\#05 Wang et al.~\cite{Wang2023RethinkingMR}    &ICML 2023  & 137 & Normal , organs and an other
 & Co-occurrence & Machine &\ding{55} \\
\hline 
\#06 KGAE~\cite{liu2021auto} &NIPS 2021  & 200  & \makecell[c]{Clinical abnormalities, \\ clinical normal} & \makecell[c]{The co-occurrence probability \\ forms weighted edges} & Machine &\ding{55} \\
\hline 
\#07 RECAP~\cite{hou2023recap}    &EMNLP 2023  & 14+ & \makecell[c]{Observation item, Spatial entity, \\ Temporal entity} &  \makecell[c]{Stable, Better, Worse, \\ spatial relation, temporal linking} & Machine &\ding{55}  \\
\hline 
\#08 ATAG~\cite{yan2023attributed}    &IEEE TMI 2023 & \makecell[c]{IU-Xray: 41 abnormal+106 attribute; \\MIMIC-CXR : 47 abnormal+209 attribute} & Anomalous, Attribute 
& \makecell[c]{3 (suggestive of, modify, located at) \\ and Co-occurrence} & Machine &\ding{55} \\
\hline 
\#09 M3KG   &Ours  & \makecell[c]{2477 entities \\ + 6943 vision tokens} & \makecell[c]{Anatomy, Disorder, Concept, \\ Device, Procedure, Size, vision tokens}  & 3 (suggestive of, modify, located at) & Machine & \checkmark \\
\hline \toprule [0.5 pt]
\end{tabular}}
\end{table*}

Despite significant improvements, these models are still limited by: 
1). Relying solely on manually annotated knowledge graphs is limited in scale; meanwhile, large language models have been exposed to vast amounts of data, making the guidance provided by existing knowledge graphs relatively limited. 
2). Existing medical knowledge graphs primarily focus on semantic-level representations, while neglecting the contribution of multi-modal information such as images. A single image can be worth a thousand words for certain diseases. 
3). Existing models adopt fixed knowledge graphs, but different cases require varying levels of detail, such static graphs struggle to support multi-level knowledge associations from macroscopic to microscopic scales. 
Therefore, it is natural to raise the following question: \textit{How can we design an accurate, large-scale, multi-modal medical knowledge graph and guide the large language models for high-performance medical report generation?}

\begin{figure*}
\centering
\includegraphics[width=\linewidth]{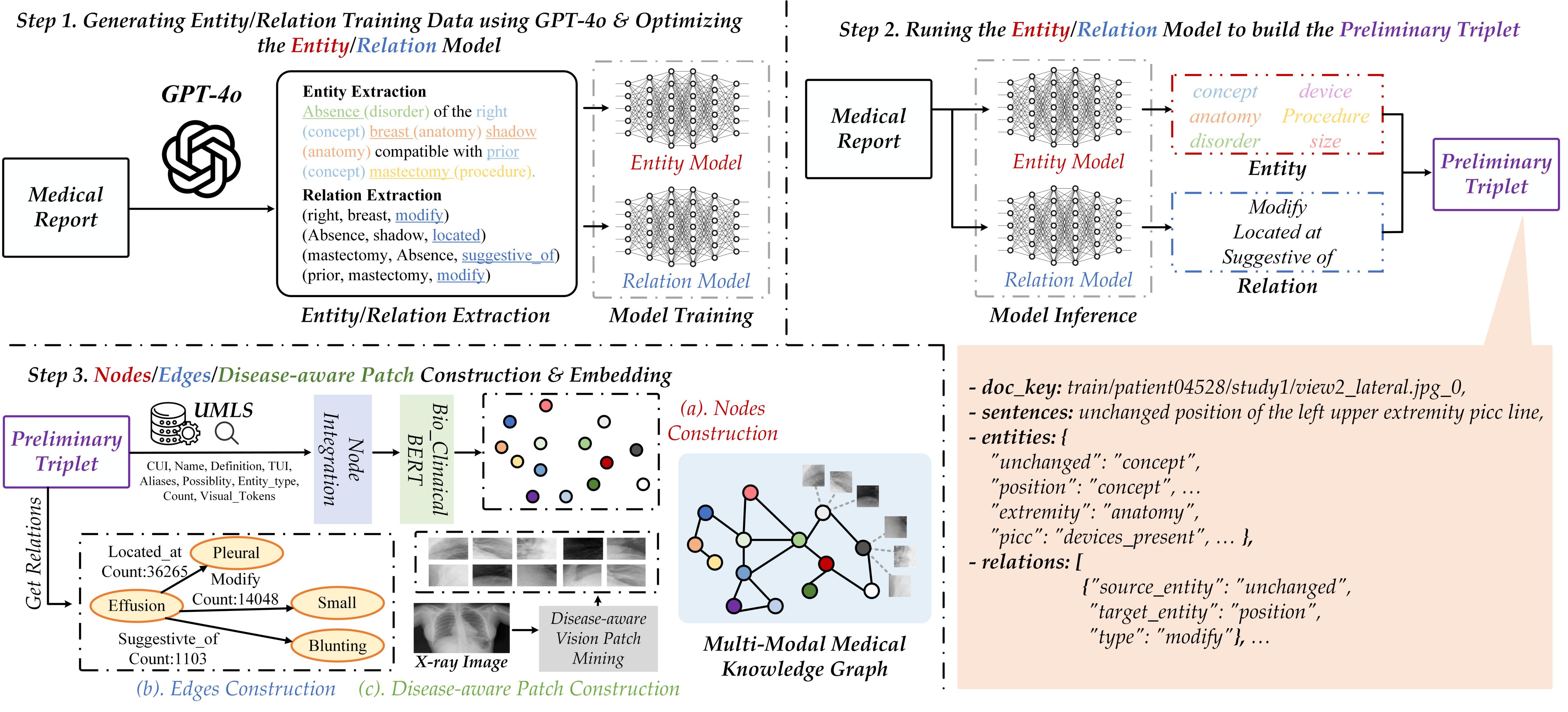}
\caption{An illustration of the proposed multi-modal medical knowledge graph M3KG.} 
\label{fig:KG_construction}
\end{figure*}

\begin{figure*}[!htp]
\centering
\includegraphics[width=0.9\linewidth]{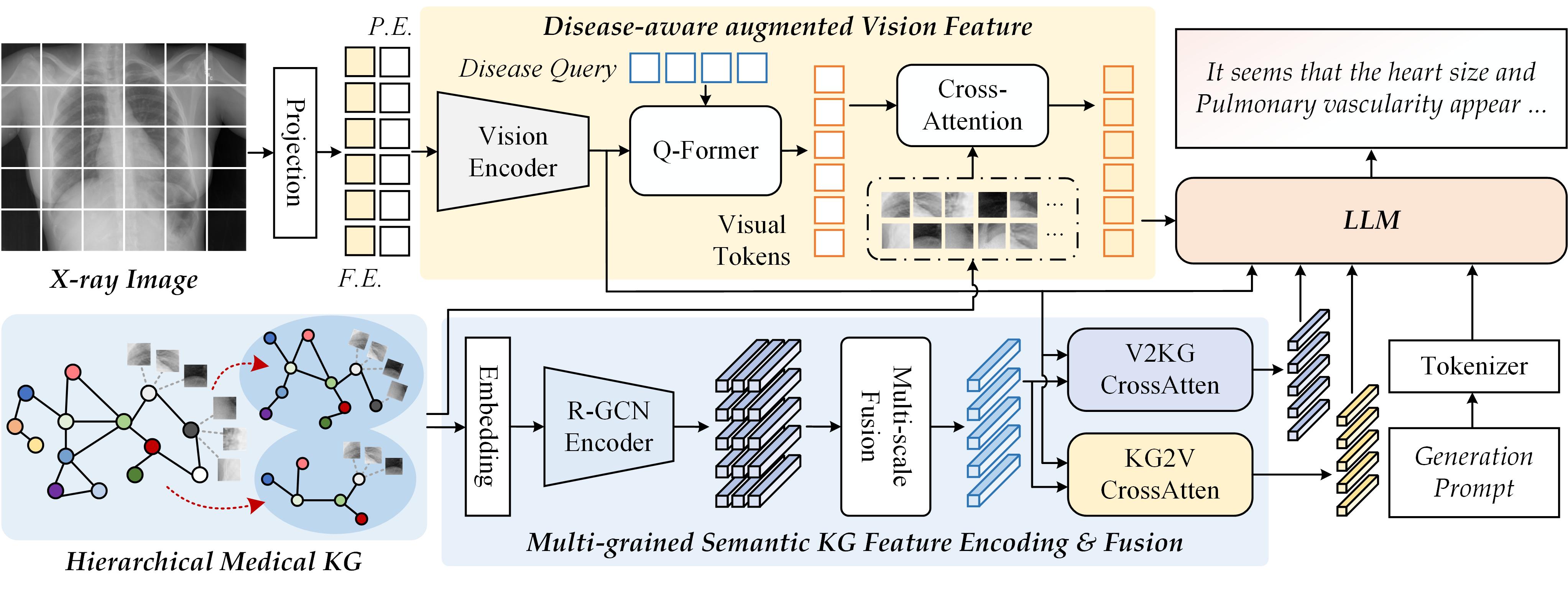}
\caption{An overview of our proposed hierarchical knowledge graph guided X-ray medical report generation framework, termed R2GenKG. }
\label{fig:R2GenKG_framework}
\end{figure*}

In this paper, we first build a new multi-modal medical knowledge graph (KG) based on ground truth reports, which contains three main stages as shown in Fig.~\ref{fig:KG_construction}. In the first stage, we adopt the LLM GPT-4o~\cite{hurst2024gpt4o} to generate training data of entities and relations and obtain these two models. We infer the two models to build the preliminary triplet in the second stage. We extract the disease-aware vision patches, nodes, and edges to build the multi-modal medical knowledge graph in the third stage. 
Based on this knowledge graph, we further propose a medical KG augmented large language model-based report generation framework, termed R2GenKG. Given the X-ray image, we extract its features using Swin-Transformer encoder and align them with the LLM using Q-former. Then, we retrieve disease-aware vision patches from the multi-modal KG to enhance the representation learning of input image. Meanwhile, we sample the medical KG to obtain multi-grained semantic KG and encode them using R-GCN~\cite{schlichtkrull2018RGCN}. Then, we fuse them and conduct cross-attention from vision to KG and vice versa. Finally, we feed the vision tokens, KG enhanced tokens, and generate a prompt into the LLM to generate the medical report. An overview of our framework can be found in Fig.~\ref{fig:R2GenKG_framework}.

To sum up, the contributions of this paper can be summarized as the following three aspects: 

1). We propose a new multi-modal medical knowledge graph construction system, termed M3KG. It generates large-scale knowledge graphs for the widely used benchmark datasets, which builds solid foundations for KG-based medical report generation.
    
2). We propose a novel hierarchical multi-grained knowledge graph augmented LLM-based report generation framework, termed R2GenKG. It fully utilizes the multi-modal and multi-granularity information from the KG to enhance the representation of visual features, and significantly improves the model's capability for clinical disease discovery by incorporating medical knowledge.
    
3). Extensive experiments on multiple benchmark datasets for medical report generation fully validated the effectiveness of our proposed KG and framework.

\section{Related Works}

In this section, we will introduce the related works on the Radiology Report Generation, Knowledge Graph, and large language models. More details can be found in the following surveys~\cite{wang2023MMPTMSurvey} and paper list\footnote{\url{github.com/Event-AHU/Medical_Image_Analysis}}.

\subsection{Radiology Report Generation}

In early studies of Radiology Report Generation (RRG), CNN-LSTM~\cite{jing2017automatic, liu2021contrastive, gajbhiye2022translating} models were widely adopted for radiology report generation tasks. For instance, Jing et al.~\cite{jing2017automatic} proposed a hierarchical LSTM model to address the challenge of generating long sentences, incorporating both sentence-level and word-level LSTM networks. Liu et al~\cite{liu2021contrastive} employed ResNet-50 as the encoder and LSTM as the decoder, while introducing a contrastive attention module. 
Alfarghaly et al.~\cite{alfarghaly2021automated} utilized a Transformer-based encoder network to combine visual features with semantic text embeddings of patient demographics, synthesizing comprehensive radiology reports. 
Wang et al.~\cite{wang2023metransformer} introduced multiple learnable expert tokens for the Transformer architecture. 


With advancements in multi-modal learning and large-scale pretrained language models, report generation has achieved significant improvements in both accuracy and naturalness~\cite{wang2024pretrainXray, wang2023r2gengpt, Wang_2025_CVPR, li2023DCL, liu2021auto}. R2GenGPT~\cite{wang2023r2gengpt} implemented the Llama2-7B~\cite{touvron2023llama} model as its decoder, demonstrating exceptional performance.
MambaXray-VL~\cite{Wang_2025_CVPR} substantially enhanced report generation performance through a multi-stage pretraining strategy and established the CXPMRG-Bench benchmark to systematically evaluate 16 LLMs. 
Additionally, several studies have leveraged structured medical knowledge to facilitate the generation of high-quality diagnostic reports~\cite{li2023DCL, liu2021auto}. For example, DCL~\cite{li2023DCL} introduced a dynamic knowledge graph based on pre-constructed organ-disease maps to optimize vision-text alignment, thereby improving both the quality and comprehensiveness of generated reports.

\subsection{Knowledge Graph} 

Knowledge graphs represent structured knowledge bases that can capture intrinsic relationships between diseases and organs in Radiology Report Generation (RRG) tasks, which can be integrated into the report generation process to enhance model performance. Li et al.~\cite{li2025context} employed Graph Convolutional Networks (GCNs) to model medical knowledge graphs, capturing relationships between different diseases. Jain et al.~\cite{Jain2021RadGraphEC} proposed RadGraph, a large-scale annotated dataset containing clinical entities and their relationships, providing substantial data support for research. Yan et al.~\cite{yan2022memory} developed a Memory-Aligned Knowledge Graph (MaKG) framework that aligns abnormal features in medical images with semantic information in knowledge graphs. Wang et al.~\cite{Wang2023RethinkingMR} constructed a comprehensive knowledge graph encompassing 137 disease types to reveal disease relationships, effectively addressing the ``long-tail" problem in datasets and improving the accuracy of rare disease descriptions. Yan et al.~\cite{yan2023attributed} introduced an Automatic Tagging and Attribute Graph (ATAG) structure that automatically constructs fine-grained abnormality graphs to capture detailed pathological characteristics. Liang~\cite{liang2024divide} proposed a Divide-and-Conquer approach that distinguishes between normal and abnormal attributes within knowledge graphs. Hou et al.~\cite{hou2025radarenhancingradiologyreport} presented the RADAR framework, which combines internal knowledge from Large Language Models (LLMs) with externally retrieved knowledge to reduce redundant information. 

\section{Methodology}

\subsection{Overview} 
We first introduce the construction process of the multimodal knowledge graph. This knowledge graph employs RGCN~\cite{schlichtkrull2018RGCN} to extract features at each scale, achieving multi-granularity knowledge coverage by loading medical knowledge graph data at five different scales. For visual information processing, Q-former~\cite{li2023blip} focuses on key information within the images and utilizes the Cross-attention~\cite{chen2021crossvit} mechanisms to interact with disease features, querying the visual components in the graph to retrieve visual knowledge. Furthermore, to facilitate effective transformation between graph and image modalities, we introduce two cross-attention modules: KG2V (Knowledge Graph to Vision) and V2KG (Vision to Knowledge Graph). Finally, these features are input into a large language model to generate diagnostic reports that better conform to medical standards.

\subsection{Multi-modal Knowledge Graph Construction}  
\indent
To address the problems of inefficient manual annotation in traditional medical knowledge graph construction and isolated entities with missing attribute information in automatic construction processes, 
we use ReXKG~\cite{zhang2024uncovering}, a comprehensive and trustworthy radiology knowledge graph that not only enables automated extraction of medical concepts and relations but also integrates with standardized medical knowledge bases (e.g., UMLS). By merging entities with high semantic similarity, ReXKG reduces redundancy and enhances graph coherence. The graph comprises six entity types (e.g., Anatomy, Disorder, Concept, etc.) and three relation types: \textit{modify}, \textit{located at}, and \textit{suggestive of}. Initially, a model such as GPT-4o is used to annotate a subset of radiology reports. These annotations subsequently serve to train a Named Entity Recognition (NER) model and a relation extraction model based on Princeton’s PURE framework.


Each entity is associated with a CUI, along with attributes such as \textit{Aliases}, \textit{Definition}, and \textit{entity\_type}. For example, the entity “Lung” is assigned the CUI \textit{C0024109}, has aliases such as \textit{lung structure} and \textit{pulmones}, an entity type of \textit{Anatomy}, and a definition: “Either of the pair of organs occupying the cavity of the thorax that effect the aeration of the blood.” The knowledge graph is constructed using relation triples in the format \{head entity, tail entity, relation\}, such as \{effusion, pleural, located\_at\}.
For the visual feature component of the graph, we adopt the method described in AM-MRG~\cite{wang2025AMMRG}. Specifically, GradCAM~\cite{jacob2021pytorch}is used to generate an activation map \( M \in \mathbb{R}^{H \times W} \) to extract disease-related visual features, where a threshold \( \tau \) is applied to identify regions of interest. The resulting disease visual features correspond to the 14 classification labels defined in the CheXpert\_plus dataset.

\subsection{Input Encoding Networks}  
\indent
To construct the input representation for each node extracted from the knowledge graph built from medical reports, we concatenate the node's attribute values into a complete input sequence as follows:
\[
\text{text}_i = \text{Concat}(\text{CUI}_i, \text{Name}_i, \text{Definition}_i, \text{TUI}_i, \text{Aliases}_i, \ldots)
\]

We utilize Bio\_ClinicalBERT~\cite{alsentzer2019publicly}, a pre-trained language model optimized for clinical and biomedical text, to encode the textual input. The model produces contextual embeddings for each token in the sequence:
\[
H_i = \mathrm{BERT}(\text{text}_i) \in \mathbb{R}^{L_i \times 768}
\]
where \( L_i \) is the length of the input sequence \( \text{text}_i \), and \( H_i \) is the sequence of hidden states. To obtain a fixed-size representation for each node, we apply mean pooling over all token embeddings. This results in a node embedding matrix \( V\in \mathbb{R}^{n \times d} \), where \( n \) is the number of nodes and \( d = 768 \) is the dimensionality of the BERT hidden layer.

The relationships between entity nodes involve multiple types of relations, thus we adopt RGCN~\cite{schlichtkrull2018RGCN} to handle such graph data. By incorporating relation types, each edge can carry distinct semantics or categories. We process the triples to construct the edge connection matrix \( \text{$edge_{index}$} \in \mathbb{R}^{2 \times n_r} \), which defines the edge connections, and the edge type matrix \( \text{$edge_{type}$} \in \mathbb{R}^{1 \times n_r} \), which represents the edge types, where \( n_r \) denotes the number of relationships between nodes.

\begin{itemize}
  \item \textbf{Edge Index}:
  \[
  \mathrm{\text{$edge_{index}$}} = \begin{bmatrix}
  h_1 & h_2 & \cdots & h_{n_r} \\
  t_1 & t_2 & \cdots & t_{n_r}
  \end{bmatrix}
  \]
  
  \item \textbf{Edge Type }:
  \[
  \mathrm{\text{$edge_{type}$}} = [r_1, r_2, \ldots, r_{n_r}]
  \]
\end{itemize}
where \( r_i \) is an integer encoding of a relation type (e.g., located\_at = 0, modify= 1).

We input the node embedding matrix \text{$edge_{index}$} and \text{$edge_{type}$} into a two-layer RGCN to propagate relational information across the graph. The RGCN update rule for the representation of node \( i \) at the \( l \)-th layer is defined as:
\[
h_i^{(l)} = \sigma \left( \sum_{r \in \mathcal{R}} \sum_{j \in \mathcal{N}_i^r} \frac{1}{c_{i, r}} W_r^{(l)} h_j^{(l-1)} + W_0^{(l)} h_i^{(l-1)} \right)
\]
where 
\( \mathcal{R} \) is the set of all relation types, \( \mathcal{N}_i^r \) is the set of neighbors of node \( i \) under relation \( r \), \( c_{i, r} \) is a normalization constant (e.g., number of neighbors),  \( W_r^{(l)} \)is the trainable weight matrix for relation \( r \) at layer \( l \), \( W_0^{(l)} \)is the self-loop transformation matrix, and \( \sigma \)is a non-linear activation function, such as ReLU.

\subsection{Hierarchical Knowledge Injection based LLM} 
In practice, the frequencies of relationships between entity nodes vary significantly, some occur tens of thousands of times, while others only once or twice. To enable the model to learn more stable patterns and reduce computational overhead, we sort the relationships in descending order of frequency and retain only the frequently occurring ones along with their associated entity nodes.

Based on the number of nodes, we construct knowledge graphs of multiple scales. Subgraphs with fewer nodes represent coarse-grained structures, facilitating global understanding, while those with more nodes offer fine-grained representations, preserving local details. This design allows for both global and local comprehension of the graph structure. The coarse subgraphs help filter out redundant information, while the fine-grained graphs retain more detailed semantics, enhancing the representational capacity of the graph. Finally, all scales are fused into a unified graph.

To integrate information across different scales, we introduce a multi-scale fusion module using the self-attention mechanism, which captures hierarchical dependencies between features of different scales. For each scale $i \in \{1, 2, \dots, S\}$, the node feature matrix is denoted by $\mathbf{X}_i \in \mathbb{R}^{N_i \times D}$, where $N_i$ is the number of nodes at scale $i$ and $D$ is the feature dimension. We enhance node features with learnable scale encodings $\mathbf{E}_{\text{scale}}(s_i)$ and positional encodings $\mathbf{E}_{\text{pos}}(p_i)$:

\[
\mathbf{X}_i' = \mathbf{X}_i + \mathbf{E}_{\text{scale}}(s_i) + \mathbf{E}_{\text{pos}}(p_i)
\]

All scale-level features are concatenated into a single matrix $\mathbf{X'} \in \mathbb{R}^{N_{\text{total}} \times D}$, where $N_{\text{total}} = \sum_{i=1}^{S} N_i$. We then apply a self-attention mechanism to compute the attention weights. The input $\mathbf{X'}$ is linearly projected to obtain queries, keys, and values. The self-attention output is computed as:
\[
\text{Attention}(\mathbf{Q}, \mathbf{K}, \mathbf{V}) = \text{Softmax} \left( \frac{\mathbf{Q} \mathbf{K}^\top}{\sqrt{d_k}} \right) \mathbf{V}
\]

The resulting fused feature matrix $\mathbf{X}'' \in \mathbb{R}^{N_{\text{total}} \times D}$ is then segmented back into each scale’s output via slicing: $\mathbf{X}_i'' = \mathbf{X}''[\text{start}_i: \text{end}_i]$. We use the output from the 300-node scale as the final graph representation.

Given a chest X-ray image $I$, we extract visual features $\mathbf{F}_v$ using a pretrained Swin Transformer~\cite{liu2021swin}. The image is then passed through a Q-Former~\cite{li2023blip} to generate query features $\mathbf{Q} \in \mathbb{R}^{C \times D}$, where $C = 14$ represents 14 visual concepts. The disease visual knowledge graph is denoted as $\mathbf{K}_V \in \mathbb{R}^{N_v \times D}$. We use cross-attention to retrieve relevant visual knowledge features $\mathbf{F}_{kv}$:
\[
F_{kv} = \text{CrossAttention}(\mathbf{F}_v,\mathbf{K}_v, \mathbf{K}_v)
\]
We further enhance interaction between the visual features $\mathbf{F}_v$ and graph features $\mathbf{X}''$ using a KG2V Crossattention and V2KG Crossattention, yielding features $\mathbf{F}_{n2g}$ and $\mathbf{F}_{g2n}$.

Finally, these features are projected into the Llama2~\cite{touvron2023llama} embedding space and concatenated as:
\[
\mathbf{F} = \text{Concat}(\mathbf{F}_v, \mathbf{F}_{kv}, \mathbf{F}_{n2g}, \mathbf{F}_{g2n}) \in \mathbb{R}^{n_f \times 4096}
\]
The final representation $\mathbf{F}$ is used as input to Llama2~\cite{touvron2023llama} for report generation.

\subsection{Loss Function} 
All modules are trained end-to-end via backpropagation to fully exploit the model’s potential in graph reasoning and visual query representation.
We use Cross-Entropy Loss as the objective function for the generation task, aiming to minimize the discrepancy between the generated text and the ground-truth descriptions:
\[
\mathcal{L}_{gen} = - \sum_{t=1}^{T} \log P(y_t \mid y_{<t}, \mathbf{F}, \mathbf{T}_{\text{prompt}})
\]
where $y_t$ is the ground-truth label of the sample at time step $t$, and $\mathbf{F}$ represents the fused feature representation from multiple sources, $\mathbf{T}_{\text{prompt}}$ denotes the tokenized generation prompt.

\begin{table*}[]
\caption{Comparison of our model’s performance on IU X-ray and Chexpert plus datasets. The best result is highlighted in bold.}
\label{tab:results_iu_chexpertplus}
\resizebox{\linewidth}{!}{
\begin{tabular}{c|l|l|ccccccc}
\hline \toprule [0.5 pt] 
\textbf{Dataset} & \textbf{Methods} & \textbf{Publication} & \textbf{BLEU-1} & \textbf{BLEU-2} & \textbf{BLEU-3} & \textbf{BLEU-4} & \textbf{ROUGE-L} & \textbf{METEOR} & \textbf{CIDEr} \\ 
\hline
\multirow{13}{*}{\textbf{IU X-Ray}} 
 & R2Gen~\cite{chen2020generating} & EMNLP 2020 & 0.470 & 0.304 & 0.219 & 0.165 & 0.371 & 0.187 & - \\
& SentSAT+KG~\cite{zhang2020radiology} & AAAI 2020 & 0.441 & 0.291 & 0.203 & 0.147 & 0.367 & - & - \\
 & R2GenCMN~\cite{chen2022crossmodalmemorynetworksradiology} & ACL-IJCNLP 2021 & 0.475 & 0.309 & 0.222 & 0.170 & 0.375 & 0.191 & - \\
 & PPKED~\cite{liu2021exploring} & CVPR 2021 & 0.483 & 0.315 & 0.224 & 0.168 & 0.376 & 0.187 & 0.351 \\
 & AlignTrans~\cite{you2021aligntransformer} & MICCAI 2021 & 0.484 & 0.313 & 0.225 & 0.173 & 0.379 & 0.204 & - \\
 & CMCL~\cite{liu2022competence} & ACL 2021 &0.473 & 0.305 & 0.217 & 0.162 & 0.378 & 0.186 & - \\
 & DCL~\cite{li2023DCL} & CVPR 2023 & - & - & - & 0.163 & 0.383 & 0.193 & 0.586 \\
 & R2GenGPT~\cite{wang2023r2gengpt} & Meta Radiology 2023 & 0.465 & 0.299 & 0.214 & 0.161 & 0.376 & 0.219 & 0.542 \\
 & PromptMRG~\cite{jin2024promptmrg} & AAAI 2024 & 0.401 & - & - & 0.098 & 0.160 & \textbf{0.281} & - \\ 
 & SILC~\cite{liu2024multi} & IEEE TMI 2024 & 0.472 & 0.321 & \textbf{0.234} & 0.175 & 0.379 & 0.192 & 0.368 \\
 & DuCo-Net~\cite{rahman2025duco} & IEEE Access 2025 & \textbf{0.500} & \textbf{0.330} & 0.220 & 0.160 & 0.260 & 0.240 & - \\
 \cline{2-10} 

 & R2GenKG &Ours & 0.468 & 0.312 & 0.231 &\textbf{0.181} & \textbf{0.383} & 0.218 & \textbf{0.701} \\
 \hline \toprule [0.5 pt] 
\multirow{13}{*}{\textbf{CheXpert Plus}} 
 & R2Gen~\cite{chen2020generating} & EMNLP 2020 & 0.301 & 0.179 & 0.118 & 0.081 & 0.246 & 0.113 & 0.077 \\
 & R2GenCMN~\cite{chen2022crossmodalmemorynetworksradiology} & ACL-IJCNLP 2021 & 0.321 & 0.195 & 0.128 & 0.087 & 0.256 & 0.127 & 0.102 \\
 & XProNet~\cite{wang2022cross} & ECCV 2022 & 0.364 & 0.225 & 0.148 & 0.100 & 0.265 & 0.146 & 0.121 \\
 & ORGan~\cite{hou2023organ} & ACL 2023 & 0.320 & 0.196 & 0.128 & 0.086 & 0.261 & 0.135 & 0.107 \\
 & R2GenGPT~\cite{wang2023r2gengpt} & Meta Radiology 2023 & 0.361 & 0.224 & 0.149 & 0.101 & 0.266 & 0.145 & 0.123 \\
 & ASGMD~\cite{xue2024generating} & ESWA 2024 & 0.267 & 0.149 & 0.094 & 0.063 & 0.220 & 0.094 & 0.044 \\
 & Token-Mixer~\cite{yang2024token} & IEEE TMI 2024 & 0.378 & 0.231 & 0.153 & 0.091 & 0.262 & 0.135 & 0.098 \\
 & PromptMRG~\cite{jin2024promptmrg} & AAAI 2024 & 0.326 & 0.174 & - & 0.095 & 0.222 & 0.121 & 0.044 \\
 & R2GenCSR~\cite{wang2024r2gencsr} & arXiv 2024 & 0.364 & 0.225 & 0.148 & 0.100 & 0.265 & 0.146 & 0.121 \\
 & MCA-RG~\cite{xing2025mca} & MICCAI 2025 &0.367 & 0.218 & 0.149 & 0.102 & 0.266 & 0.147 & - \\
 \cline{2-10} 
 & R2GenKG &Ours & \textbf{0.376}  & \textbf{0.234} & \textbf{0.155} & \textbf{0.106} & \textbf{0.269} & \textbf{0.151} & \textbf{0.125} \\
 \hline \toprule [0.5 pt] 
\end{tabular}
}
\end{table*}

\section{Experiments} 

\subsection{Datasets and Evaluation Metric}  

In our experiments, we adopt two widely used benchmark datasets for the medical X-ray report generation, i.e., the IU-Xray~\cite{demner2016iuxray} dataset and CheXpert Plus~\cite{chambon2024CheXpertPLUS} dataset. A more detailed introduction to these datasets can be found in our supplementary material. To evaluate our R2GenKG model, we use widely adopted natural language generation (NLG) metrics, including BLEU~\cite{papineni2002bleu}, ROUGE-L~\cite{lin2004rouge}, METEOR~\cite{banerjee2005meteor}, and CIDEr~\cite{vedantam2015cider}. Additionally, we follow R2Gen~\cite{chen2020generating} and use the CE metric to assess clinical accuracy. 

\noindent $\bullet$ \textbf{IU-Xray Dataset.} 
The IU X-ray dataset is a publicly available medical imaging dataset that primarily contains chest X-ray images and their corresponding reports. The dataset, provided by the Indiana University School of Medicine, includes 7,470 chest X-ray images and 3,955 corresponding reports. Following R2Gen~\cite{chen2020generating}, R2GenGPT~\cite{wang2023r2gengpt}, we divide the dataset into training, testing, and validation sets with a ratio of 7:1:2.

\noindent $\bullet$ \textbf{CheXpert Plus Dataset.}
The CheXpert Plus dataset integrates both text and image data, aiming to enhance the performance, robustness, and fairness of machine learning models in the field of radiology. It contains 223,228 chest X-ray images and corresponding reports, covering annotations for 14 different chest pathologies, further improving the data quality. The dataset is widely used in tasks such as image diagnosis, image labeling, and report generation. To ensure fairness, we adopt the dataset partition strategy proposed in CXPMRG-Bench~\cite{Wang_2025_CVPR}.

Specifically, BLEU calculates scores based on n-gram precision to assess the similarity between generated text and reference text. ROUGE-L measures the longest common subsequence (LCS) between the generated report and the reference report. METEOR improves upon BLEU by penalizing word order inconsistencies and morphological variations. CIDEr calculates n-gram matching based on TF-IDF weights, avoiding score distortion due to biases from a single reference text. 

Specifically, we evaluate the model’s ability to correctly identify diseases, lesions, and other features using Precision, Recall, and F1 Score. High precision indicates fewer false positives, while high recall indicates the model's ability to capture most of the positive instances. F1 Score provides a more comprehensive performance evaluation by considering both precision and recall.

\subsection{Comparison on Public Benchmark Datasets}  

\noindent $\bullet$ \textbf{Analysis of the NLG Metrics.} 
To comprehensively evaluate the effectiveness of our proposed method for medical image report generation, we conducted comparative experiments on two widely used public benchmark datasets: the IU X-Ray dataset and the CheXpert Plus dataset. Table~\ref{tab:results_iu_chexpertplus} summarizes the performance of various methods based on commonly adopted natural language generation evaluation metrics, including BLEU, ROUGE-L, METEOR, and CIDEr. On the IU X-Ray dataset, our method achieved top-tier performance across multiple metrics, with BLEU-1 at 0.468, BLEU-2 at 0.312, BLEU-3 at 0.231, BLEU-4 at 0.181, ROUGE-L at 0.383, METEOR at 0.218, and a CIDEr score of 0.701. These results indicate that our approach effectively captures medical content and its linguistic structure on this relatively small-scale dataset.
On the more complex CheXpert Plus dataset, our method continued to demonstrate strong performance, achieving BLEU-1 of 0.376, BLEU-2 of 0.234, BLEU-3 of 0.155, BLEU-4 of 0.106, ROUGE-L of 0.269, METEOR of 0.151, and a CIDEr score of 0.125. Our method outperformed all baseline approaches across all metrics, highlighting superior semantic retention and linguistic coherence. Overall, our approach exhibited balanced and stable performance across both datasets, which presents a feasible and efficient solution for automatic medical image report generation.

\noindent $\bullet$ \textbf{Analysis of CE Metric.}
As shown in the Table~\ref{tab:ce_metrics}, we compared multiple medical report generation models on the CheXpert Plus dataset using Clinical Efficacy (CE) metrics, including ORGan~\cite{hou2023organ}, which also employs a knowledge graph, to evaluate the models' accuracy in identifying clinical abnormalities. The metrics used include Precision, Recall, and F1 score. It can be observed that our method outperforms all others in the two CE metrics. Although ORGan~\cite{hou2023organ} achieves a Recall of 0.287, both its Precision and F1 scores are significantly lower than those of our model. This demonstrates that our method not only ensures precision in the generated reports but also maintains a high recall, achieving a good balance. The CE metrics indicate that our model effectively identifies pathological features in medical images, ensuring its capability to recognize key clinical information.

\subsection{Implementation Details}  
For the input chest X-ray images, we use a pre-trained Swin Transformer~\cite{liu2021swin} as the visual encoder. To enable cross-attention with visual features, we employ gcn\_proj to map the graph node feature dimensions to the visual feature dimension of 1024. Then, a simple projection layer is used to map the features to the LLM feature dimension of 4096, which is concatenated with the visual knowledge and fed into the LLM embedding space for report generation. The LLM used is Llama2-7B~\cite{touvron2023llama}, and the node feature encoding is performed using Bio ClinicalBERT~\cite{alsentzer2019publicly}. After multi-scale fusion, the number of nodes used is 319, and 500 visual disease features are employed. We set the learning rate to 9e-5 and trained the model using the ADAMW~\cite{loshchilov2017decoupled} optimizer. In our experiments, the model was developed using PyTorch~\cite{paszke2019pytorch} and trained and tested on a server equipped with an NVIDIA A800SXM4-80GB GPU. More details can be found in our source code. 

\begin{table*}[ht]
\centering
\caption{Ablation study on CheXpert Plus dataset, assessing the impact of key components: RGCN (RG), multi-scale feature Fusion (MF) and disease visual graph (DVG). A “\checkmark” indicates the presence of each component, while “-” denotes its absence.}
\label{tab:ce_chexpertplus}
\begin{tabular}{ c|c|c c c|c c c }
\hline \toprule [0.5 pt] 
\textbf{Dataset} & \textbf{Setting} & \textbf{RG} & \textbf{MF} & \textbf{DVG}   
& \textbf{Precision} & \textbf{Recall} & \textbf{F1}  \\
\hline
\multirow{5}{*}{CheXpert Plus} 
& BASE & - & - & - & 0.315 & 0.224 & 0.260  \\
& (a)  & - & - & \checkmark &  0.330  & 0.251 & 0.262  \\
& (b)  & \checkmark & - & - & \textbf{0.346} & 0.273 & 0.287  \\
& (c)  & \checkmark & \checkmark & - & 0.334 & 0.273 & 0.286  \\
\cline{2-8}
& (d)  & \checkmark & \checkmark & \checkmark & 0.338 & \textbf{0.275} & \textbf{0.292} \\
\hline \toprule [0.5 pt] 
\end{tabular}
\end{table*}

\begin{table*}[ht]
\centering
\small 
\caption{Ablation study on CheXpert Plus dataset, assessing the impact of key components: RGCN (RG), multi-scale feature Fusion (MF) and disease visual graph (DVG). A “\checkmark” indicates the presence of each component, while “-” denotes its absence.}
\label{tab:ablation_nlg_chexpertplus}
\begin{tabular}{ c|c|c c c|c c c c c c c }
\hline \toprule [0.5 pt] 
\textbf{Dataset} & \textbf{Setting} & \textbf{RG} & \textbf{MF} & \textbf{DVG}   
& \textbf{BLEU-1} & \textbf{BLEU-2} & \textbf{BLEU-3} & \textbf{BLEU-4} 
& \textbf{RG-L} & \textbf{METEOR} & \textbf{CIDEr} \\
\hline
\multirow{5}{*}{CheXpert Plus} 
& BASE & - & - & - & 0.361 & 0.224 & 0.149 & 0.101 & 0.266 & 0.145 & 0.123\\
& (a)  & - & - & \checkmark &  0.368  & 0.228 & 0.151 & 0.103 & 0.267 & 0.149 &0.127\\
& (b)  & \checkmark & - & - & 0.367 & 0.229 & 0.152 & 0.104 & 0.269 & 0.147 & \textbf{0.134}\\
& (c)  & \checkmark & \checkmark & - & 0.374 & 0.232 & 0.154 & 0.105 & 0.268 & 0.150 & 0.120\\
\cline{2-12}
& (d)  & \checkmark & \checkmark & \checkmark & \textbf{0.376} & \textbf{0.234} & \textbf{0.155} & \textbf{0.106} & \textbf{0.269} & \textbf{0.151} & 0.125 \\
\hline \toprule [0.5 pt] 
\end{tabular}
\end{table*}

\subsection{Component Analysis} 

As shown in Table~\ref{tab:ce_chexpertplus} and Table~\ref{tab:ablation_nlg_chexpertplus}, we conducted comprehensive ablation studies on the CheXpert Plus dataset to evaluate the contributions of the three key components in our proposed model: the Relational Graph Convolutional Network module (RGCN, denoted as RG), the Multi-scale Feature Fusion module (MF), and the Disease Visual Graph module (DVG). The purpose of these experiments was to verify the impact of each module on the overall performance of the model. The results demonstrate that, compared to the baseline model (BASE), introducing any individual module or a combination of modules consistently led to performance improvements across both Natural Language Generation (NLG) metrics and Clinical Efficacy (CE) metrics. This indicates that each component contributes positively to enhancing the model’s representation capability and generation quality. Specifically, for NLG metrics, the BLEU-1, BLEU-2, BLEU-3, BLEU-4, ROUGE-L, METEOR, and CIDEr scores improved from 0.361, 0.224, 0.149, 0.101, 0.266, 0.145, and 0.123, respectively, to 0.376, 0.234, 0.155, 0.106, 0.269, 0.151, and 0.125. These improvements suggest that the model achieves better accuracy, fluency, and diversity in language generation, producing reports that more closely resemble authentic clinical descriptions. Regarding CE metrics, the Precision, Recall, and F1 scores reached 0.338, 0.275, and 0.292, respectively, which are significantly higher than the baseline values of 0.315, 0.224, and 0.260. This indicates that the integration of graph-structured modeling and multi-scale feature fusion also enhances the clinical semantic validity and accuracy of the generated content.

\begin{table}[ht]
\centering
\small 
\caption{A comparison of the clinical efficacy (CE) metrics between our proposed framework (Ours) and state-of-the-art methods using F1 score, precision, and recall on the CheXpert Plus dataset.}
\label{tab:ce_metrics}
\begin{tabular}{l|ccc}
\hline \toprule [0.5 pt] 
\textbf{Model} & \textbf{F1} & \textbf{Precision} & \textbf{Recall} \\
\hline
R2Gen~\cite{chen2020generating} & 0.181 & 0.318 & 0.200 \\
R2GenCMN~\cite{chen2022crossmodalmemorynetworksradiology} & 0.231 & 0.329 & 0.241 \\
WCL~\cite{yan2021weakly}& 0.256 & 0.335 & 0.259 \\
PromptMRG~\cite{jin2024promptmrg} & 0.281 & 0.258 & 0.265 \\
R2GenGPT ~\cite{wang2023r2gengpt} & 0.260 & 0.315 & 0.224 \\
ORGan~\cite{hou2023organ} & 0.277 & 0.288 & \textbf{0.287} \\
Token-Mixer~\cite{yang2024token} & 0.288 & 0.309 & 0.270 \\
\cline{1-4} 
R2GenKG (Ours) & \textbf{0.292} & \textbf{0.338} & 0.275 \\
\hline \toprule [0.5 pt] 
\end{tabular}
\end{table}

\begin{table}
\centering
\small 
\caption{Compare the effects of different KG encoders.}
\label{tab:report_ablation_KG encoders}
\begin{tabular}{c c c c c}
\hline \toprule [0.5 pt] 
\textbf{Encoder} & \textbf{BLEU-4} & \textbf{ROUGE-L} & \textbf{METEOR} & \textbf{CIDEr} \\
\hline
GCN   & 0.102 & 0.263 & 0.147  & 0.116 \\
RGCN  & \textbf{0.106} & \textbf{0.269}  & \textbf{0.151} & \textbf{0.125} \\
GAT  & 0.103 & 0.265 & 0.148 & 0.121 \\
\hline  \toprule [0.5 pt]
\end{tabular}
\end{table}

\begin{table}[t]
\centering
\caption{Efficiency analysis of R2GenKG}
\small 
\label{tab:R2GenKG_efficiency}
\begin{tabular}{l|ccc}
\hline
\textbf{Metric} & \textbf{Parameters} & \textbf{Memory Usage} & \textbf{Speed} \\
\hline 
R2GenKG & 239M & 915.63MB & 33.29s/iter \\
\hline
\end{tabular}
\end{table}

\begin{table}[ht]
\centering
\small 
\caption{Comparison of different numbers of visual features.}
\label{tab:num_ablation}
\begin{tabular}{c c c c c}
\hline \toprule [0.5 pt] 
\textbf{Number} & \textbf{BLEU-4} & \textbf{ROUGE-L} & \textbf{METEOR} & \textbf{CIDEr} \\
\hline
100   &  0.105 & 0.267 & 0.149 & 0.121 \\
300  & 0.104 & 0.267 & 0.149 & 0.122 \\
500  & \textbf{0.106} & \textbf{0.269}  & \textbf{0.151} & \textbf{0.125} \\
700  & 0.103  & 0.266 & 0.148 &  0.120 \\
1000  & 0.105 & 0.268 & 0.149 & 0.124 \\
\hline \toprule [0.5 pt] 
\end{tabular}
\end{table}

\begin{table}[tb!]
\centering
\small 
\caption{Compare the effects of different numbers of entities.}
\label{tab:entity_ablation}
\begin{tabular}{c c c c c}
\hline \toprule [0.5 pt] 
\textbf{\#Entity} & \textbf{BLEU-4} & \textbf{ROUGE-L} & \textbf{METEOR} & \textbf{CIDEr} \\
\hline
100   & 0.104 & 0.266 &  0.149 &  0.124 \\
200  &  0.104 & 0.266 & 0.148 & 0.124 \\
300  & \textbf{0.106} & \textbf{0.269}  & \textbf{0.151} & \textbf{0.125} \\
400  & 0.100 & 0.259 & 0.144 & 0.109 \\
500  & 0.101 & 0.265 & 0.148 & 0.125 \\
\hline \toprule [0.5 pt] 
\end{tabular}
\end{table}

\subsection{Ablation Study} 
\noindent $\bullet$ \textbf{Analysis of Numbers of Entity.}
To evaluate the impact of the number of entity nodes on medical report generation quality, we conducted an ablation study, evaluating the model using four standard metrics: BLEU-4, ROUGE-L, METEOR, and CIDEr. The results are presented in Table~\ref{tab:entity_ablation}. As the number of entity nodes increased from 100 to 300, all evaluation metrics showed varying degrees of improvement. Notably, optimal overall performance was achieved when the number of entity nodes was set to 300, with BLEU-4 reaching 0.106, ROUGE-L at 0.269, METEOR at 0.151, and CIDEr at 0.125. Further increases to 400 and 500 nodes resulted in performance degradation across metrics. From these observations, we hypothesize that a moderate number of entity nodes effectively enhances model capabilities. However, excessive entity nodes may introduce redundant or noisy information, negatively impacting model performance. Therefore, setting the number of entity nodes to 300 achieves the optimal balance for this task.

\noindent $\bullet$ \textbf{Analysis of Numbers of Visual Features.} 
To investigate the impact of visual feature quantity on medical image report generation performance, we conducted a series of systematic ablation studies, as shown in Table ~\ref{tab:num_ablation}. When the number of visual features increased from 100 to 500, the model demonstrated consistent improvement across multiple natural language generation (NLG) metrics. Performance peaked at 500 features, achieving the highest scores across all metrics. However, further increasing the feature count to 700 and 1000 resulted in slight performance degradation. In conclusion, our method achieves optimal performance with 500 visual features, indicating that this configuration strikes an effective balance between the richness of visual information and the expressiveness of the generated reports.

\begin{figure*}[!htp]
\newcommand{\MatchingKGGpt}[1]{\sethlcolor{pink}\hl{#1}}
\newcommand{\MatchingGpt}[1]{\sethlcolor{cyan}\hl{#1}}
\newcommand{\MatchingKG}[1]{\sethlcolor{yellow}\hl{#1}}
\centering
\resizebox{\textwidth}{!}{
\begin{tabular}{lp{6.5cm}p{6.5cm}p{6.5cm}}
\hline \toprule [0.5 pt] 
\multicolumn{1}{c}{\textbf{Image}} & \multicolumn{1}{c}{\textbf{Ground Truth}} & \multicolumn{1}{c}{\textbf{Ours}} & \multicolumn{1}{c}{\textbf{R2GenGPT}} \\ 
\hline \toprule [0.5 pt] 
\begin{minipage}[b]{0.5\columnwidth}\vspace{7pt}\centering\raisebox{-.9\height}{\includegraphics[width=\linewidth]{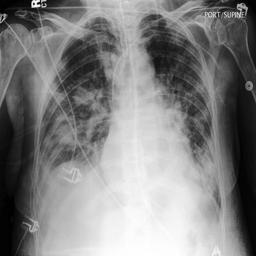}}\end{minipage}
& \MatchingKGGpt{Submitted for review is a single frontal portable view of the chest.} \MatchingKG{An endotracheal tube is seen with its tip in the trachea. A feeding tube is seen with its tip below the diaphragm. A left} subclavian \MatchingKG{venous catheter is seen with its tip in the superior vena cava.} \MatchingKGGpt{The cardiomediastinal silhouette appears unremarkable.}
& \MatchingKG{Submitted for review is a single frontal portable view of the chest. An endotracheal tube is seen with its tip in the trachea.    A feeding tube is seen with its tip below the diaphragm. A left} internal jugular \MatchingKG{venous catheter is seen with its tip in the superior vena cava. The cardiomediastinal silhouette appears unremarkable.} The lungs demonstrate diffuse alveolar opacities consistent.
& \MatchingGpt{Submitted for review is a single frontal portable view of the chest.} There has been interval placement of a left subclavian venous catheter with its tip in the superior vena cava. There is redemonstration of diffuse bilateral patchy air space opacities consistent with pulmonary edema. There is no evidence of pneumothorax. \MatchingGpt{The cardiomediastinal silhouette is unremarkable.} \\ 
\begin{minipage}[b]{0.5\columnwidth}\vspace{7pt}\centering\raisebox{-.9\height}{\includegraphics[width=\linewidth]{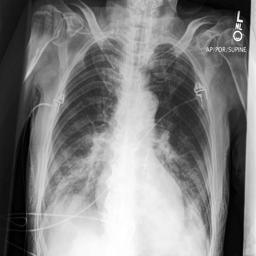}}\end{minipage} 
& \MatchingKG{Submitted for review is a single frontal portable view of the chest. An endotracheal tube is seen with its tip in the trachea. The cardiomediastinal silhouette appears unremarkable.} The lungs demonstrate confluent alveolar opacification in the bases bilaterally with blunting of the left costophrenic angle. \MatchingKG{There is diffuse increased reticular markings with indistinct pulmonary vessels} with increase in densit.
 & \MatchingKG{Submitted for review is a single frontal portable view of the chest. An endotracheal tube is seen with its tip in the trachea.} There is a nasogastric tube seen with its tip below the diaphragm. \MatchingKG{The cardiomediastinal silhouette appears unremarkable. The lungs demonstrate diffuse increased reticular markings and indistinctness of the pulmonary vasculature consistent} with pulmonary edema.
 & Single portable supine view of the chest demonstrates interval placement of an endotracheal tube with the tip at the level of the clavicles. There is a nasogastric tube with the tip in the stomach. There are diffuse patchy air space opacities bilaterally which may represent pulmonary edema. There is mild cardiomegaly.   no bony fractures or acute osseous abnormalities are seen.\\ 
\begin{minipage}[b]{0.5\columnwidth}\vspace{7pt}\centering\raisebox{-.9\height}{\includegraphics[width=\linewidth]{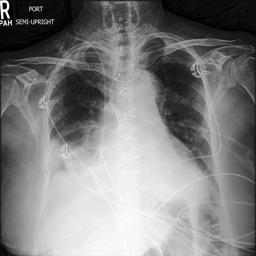}}\end{minipage} 
& Redemonstration of postsurgical changes and \MatchingKG{unchanged positioning of right internal jugular central venous catheter. Low lung volumes} \MatchingKGGpt{with bibasilar opacities} \MatchingGpt{atelectasis versus consolidation.} \MatchingKGGpt{Small bilateral pleural effusions.} Persistent \MatchingGpt{mild pulmonary edema.} \MatchingKGGpt{No pneumothorax.} \MatchingKG{Enlarged cardiomediastinal silhouette.}
 & \MatchingKG{Unchanged positioning of right internal jugular central venous catheter} median sternotomy wires bioprosthetic aortic valve epicardial pacing wires and mediastinal clips. \MatchingKG{Low lung volumes }\MatchingKGGpt{with bibasilar opacities} \MatchingKGGpt{and small bilateral pleural effusions.}   \MatchingKGGpt{No pneumothorax.} \MatchingKG{Enlarged cardiac silhouette.}
 & Stable right ij central venous catheter. Interval removal of the swan-ganz catheter. Redemonstration of median sternotomy wires and mediastinal clips. Unchanged cardiomediastinal silhouette. Persistent \MatchingKGGpt{bibasilar opacities} likely \MatchingGpt{atelectasis or consolidation.} \MatchingKGGpt{Small bilateral pleural effusions.} \MatchingGpt{Mild pulmonary edema.} \MatchingKGGpt{No pneumothorax.} \\ 
\hline \toprule [0.5 pt] 
\end{tabular}
}
\caption{X-ray images and their corresponding ground-truths, along with the output of our model and R2GenGPT model generation reports in the CheXpert Plus dataset. Matching sentences in our report are highlighted in yellow, R2GenGPT matching sentences are highlighted in cyan, and sentences matching by both models are highlighted in pink.}
\label{fig:visual_report}
\end{figure*}

\begin{figure}
\centering
\includegraphics[width=0.4\textwidth]{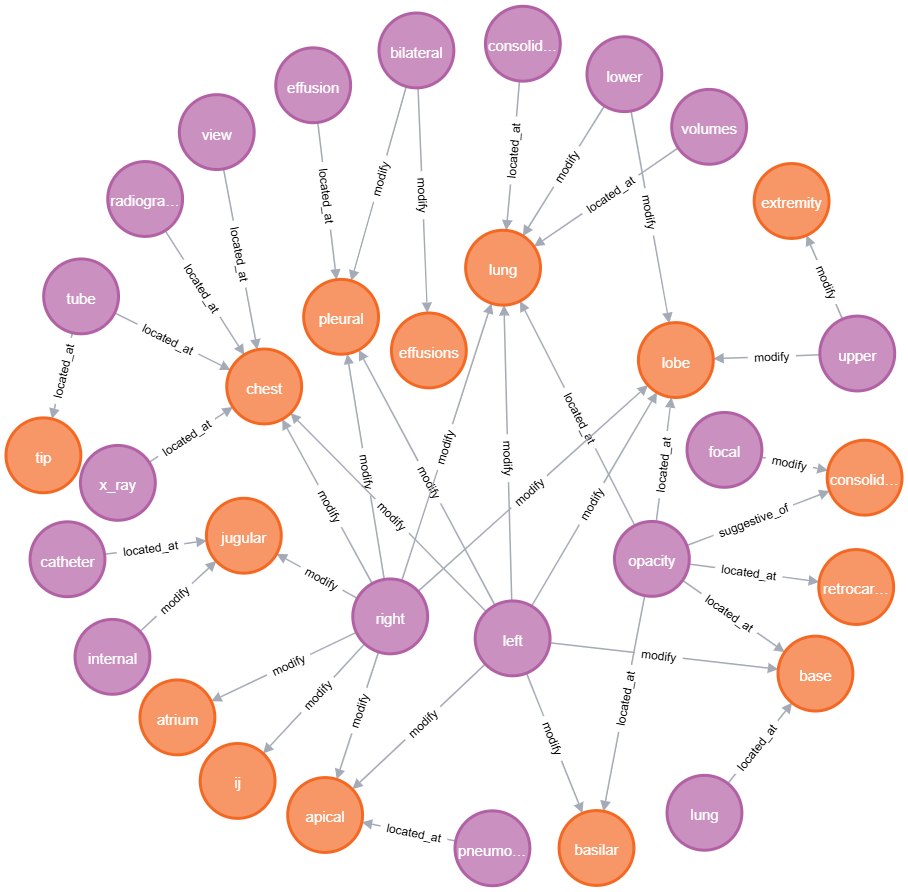}
\caption{An illustration of the part of our proposed multi-modal medical knowledge graph M3KG.}
\label{fig:KG_visualization}
\end{figure}

\noindent $\bullet$ \textbf{Analysis of Different Encoders for Knowledge Graph.}
As shown in Table~\ref{tab:report_ablation_KG encoders}, we compare three types of graph encoding methods GCN~\cite{kipf2016semi}, RGCN~\cite{schlichtkrull2018RGCN}, and GAT~\cite{velivckovic2017graph}in terms of BLEU-4, ROUGE-L, METEOR, and CIDEr metrics. The experimental results demonstrate that RGCN achieves the best performance across all evaluation metrics (e.g., BLEU-4 = 0.106, CIDEr = 0.125). We hypothesize that this is due to RGCN's ability to model multiple types of relations, which enables it to better capture the semantic dependencies among various clinical entities in the medical knowledge graph.

\subsection{Visualization}

\noindent $\bullet$ \textbf{Report Generation.~} 
As shown in Figure~\ref{fig:visual_report}, we present some examples to demonstrate the effectiveness of our proposed R2GenKG model for X-ray image-based report generation. For specific X-ray images, we compare the ground truth with the reports generated by the R2GenKG model and the R2GenGPT model. To provide a more intuitive visualization, we highlight the parts that match the ground truth: the yellow-highlighted areas represent the portions of the report generated by the R2GenKG model that align with the ground truth; the cyan-highlighted areas represent the portions of the report generated by the R2GenGPT model that align with the ground truth; and the pink-highlighted areas indicate the portions where both the R2GenKG and R2GenGPT models' reports match the ground truth. From the visualization results, it is evident that the reports generated by the R2GenKG model are of higher quality and more consistent with the ground truth compared to those generated by the R2GenGPT model.

\noindent $\bullet$ \textbf{Multi-modal Knowledge Graph.~} 
As shown in Figure~\ref{fig:KG_visualization}, we implemented the graph visualization using Neo4j. Neo4j is one of the most popular and powerful graph databases, offering advanced capabilities for graph data modeling and querying. It also provides various graphical interface tools that help users intuitively understand the structure and patterns of graph data. In this study, due to the large number of nodes in the dataset, we only display a portion of the relationships between the nodes to ensure clarity and readability of the graph. The nodes in the figure represent medical entities, and each arrow indicates a relationship, where one node performs a certain type of operation or influence on another node.

\subsection{Parameter Analysis} 
We analyze the efficiency of R2GenKG on the CheXpert Plus dataset, as shown in Table~\ref{tab:R2GenKG_efficiency}. The number of parameters in our model is 915.63MB, the test speed is 33.29s/iter, and the number of trainable parameters is 239M.

\subsection{Limitation Analysis}  
The overall framework of R2GenKG involves multiple modules, particularly the invocation of large language models, which require training and inference on high-performance GPUs. This incurs significant computational costs, limiting its deployment potential in clinical scenarios with restricted resources. Additionally, there are differences in granularity and semantic space between visual disease features and textual graphs. The current model lacks deep alignment mechanisms in terms of structural hierarchy and semantic representation, resulting in limited cross-modal fusion performance and hindering the full potential of knowledge-guided reasoning.

\section{Conclusion} 
In this paper, we propose a novel multi-scale, multimodal knowledge graph-enhanced framework, R2GenKG, aimed at improving the quality of automatic medical report generation based on X-ray images. We first construct a large-scale multimodal medical knowledge graph, M3KG, and leverage multi-granularity knowledge graph encoding, disease-aware visual tag retrieval, and cross-modal feature interaction to effectively address the limitations of existing models in clinical knowledge utilization and disease diagnosis capabilities. 
Extensive experimental results demonstrate that R2GenKG outperforms existing methods on multiple public benchmark datasets. 

{
    \small
    \bibliographystyle{ieeenat_fullname}
    \bibliography{main}

\begin{thebibliography}{61}
\providecommand{\natexlab}[1]{#1}
\providecommand{\url}[1]{\texttt{#1}}
\expandafter\ifx\csname urlstyle\endcsname\relax
  \providecommand{\doi}[1]{doi: #1}\else
  \providecommand{\doi}{doi: \begingroup \urlstyle{rm}\Url}\fi

\bibitem[Alfarghaly et~al.(2021)Alfarghaly, Khaled, Elkorany, Helal, and
  Fahmy]{alfarghaly2021automated}
Omar Alfarghaly, Rana Khaled, Abeer Elkorany, Maha Helal, and Aly Fahmy.
\newblock Automated radiology report generation using conditioned transformers.
\newblock \emph{Informatics in Medicine Unlocked}, 24:\penalty0 100557, 2021.

\bibitem[Alsentzer et~al.(2019)Alsentzer, Murphy, Boag, Weng, Jin, Naumann, and
  McDermott]{alsentzer2019publicly}
Emily Alsentzer, John~R Murphy, Willie Boag, Wei-Hung Weng, Di Jin, Tristan
  Naumann, and Matthew McDermott.
\newblock Publicly available clinical bert embeddings.
\newblock \emph{arXiv preprint arXiv:1904.03323}, 2019.

\bibitem[Banerjee and Lavie(2005)]{banerjee2005meteor}
Satanjeev Banerjee and Alon Lavie.
\newblock Meteor: An automatic metric for mt evaluation with improved
  correlation with human judgments.
\newblock In \emph{Proceedings of the acl workshop on intrinsic and extrinsic
  evaluation measures for machine translation and/or summarization}, pages
  65--72, 2005.

\bibitem[Chambon et~al.(2024)Chambon, Delbrouck, Sounack, Huang, Chen, Varma,
  Truong, Chuong, and Langlotz]{chambon2024CheXpertPLUS}
Pierre Chambon, Jean-Benoit Delbrouck, Thomas Sounack, Shih-Cheng Huang,
  Zhihong Chen, Maya Varma, Steven~QH Truong, Chu~The Chuong, and Curtis~P
  Langlotz.
\newblock Chexpert plus: Augmenting a large chest x-ray dataset with text
  radiology reports, patient demographics and additional image formats.
\newblock \emph{arXiv preprint arXiv:2405.19538}, 2024.

\bibitem[Chen et~al.(2021)Chen, Fan, and Panda]{chen2021crossvit}
Chun-Fu~Richard Chen, Quanfu Fan, and Rameswar Panda.
\newblock Crossvit: Cross-attention multi-scale vision transformer for image
  classification.
\newblock In \emph{Proceedings of the IEEE/CVF international conference on
  computer vision}, pages 357--366, 2021.

\bibitem[Chen et~al.(2020)Chen, Song, Chang, and Wan]{chen2020generating}
Zhihong Chen, Yan Song, Tsung-Hui Chang, and Xiang Wan.
\newblock Generating radiology reports via memory-driven transformer.
\newblock \emph{arXiv preprint arXiv:2010.16056}, 2020.

\bibitem[Chen et~al.(2022)Chen, Shen, Song, and
  Wan]{chen2022crossmodalmemorynetworksradiology}
Zhihong Chen, Yaling Shen, Yan Song, and Xiang Wan.
\newblock Cross-modal memory networks for radiology report generation, 2022.

\bibitem[Demner-Fushman et~al.(2016)Demner-Fushman, Kohli, Rosenman, Shooshan,
  Rodriguez, Antani, Thoma, and McDonald]{demner2016iuxray}
Dina Demner-Fushman, Marc~D Kohli, Marc~B Rosenman, Sonya~E Shooshan, Laritza
  Rodriguez, Sameer Antani, George~R Thoma, and Clement~J McDonald.
\newblock Preparing a collection of radiology examinations for distribution and
  retrieval.
\newblock \emph{Journal of the American Medical Informatics Association},
  23\penalty0 (2):\penalty0 304--310, 2016.

\bibitem[Gajbhiye et~al.(2022)Gajbhiye, Nandedkar, and
  Faye]{gajbhiye2022translating}
Gaurav~O Gajbhiye, Abhijeet~V Nandedkar, and Ibrahima Faye.
\newblock Translating medical image to radiological report: Adaptive multilevel
  multi-attention approach.
\newblock \emph{Computer Methods and Programs in Biomedicine}, 221:\penalty0
  106853, 2022.

\bibitem[Guo et~al.(2025)Guo, Yang, Zhang, Song, Zhang, Xu, Zhu, Ma, Wang, Bi,
  et~al.]{guo2025deepseekR1}
Daya Guo, Dejian Yang, Haowei Zhang, Junxiao Song, Ruoyu Zhang, Runxin Xu,
  Qihao Zhu, Shirong Ma, Peiyi Wang, Xiao Bi, et~al.
\newblock Deepseek-r1: Incentivizing reasoning capability in llms via
  reinforcement learning.
\newblock \emph{arXiv preprint arXiv:2501.12948}, 2025.

\bibitem[Hou et~al.(2023{\natexlab{a}})Hou, Cheng, Xu, Li, and
  Liu]{hou2023recap}
Wenjun Hou, Yi Cheng, Kaishuai Xu, Wenjie Li, and Jiang Liu.
\newblock Recap: Towards precise radiology report generation via dynamic
  disease progression reasoning.
\newblock \emph{arXiv preprint arXiv:2310.13864}, 2023{\natexlab{a}}.

\bibitem[Hou et~al.(2023{\natexlab{b}})Hou, Xu, Cheng, Li, and
  Liu]{hou2023organ}
Wenjun Hou, Kaishuai Xu, Yi Cheng, Wenjie Li, and Jiang Liu.
\newblock Organ: Observation-guided radiology report generation via tree
  reasoning.
\newblock \emph{arXiv preprint arXiv:2306.06466}, 2023{\natexlab{b}}.

\bibitem[Hou et~al.(2025)Hou, Cheng, Xu, Li, Hu, Li, and
  Liu]{hou2025radarenhancingradiologyreport}
Wenjun Hou, Yi Cheng, Kaishuai Xu, Heng Li, Yan Hu, Wenjie Li, and Jiang Liu.
\newblock Radar: Enhancing radiology report generation with supplementary
  knowledge injection, 2025.

\bibitem[Hurst et~al.(2024)Hurst, Lerer, Goucher, Perelman, Ramesh, Clark,
  Ostrow, Welihinda, Hayes, Radford, et~al.]{hurst2024gpt4o}
Aaron Hurst, Adam Lerer, Adam~P Goucher, Adam Perelman, Aditya Ramesh, Aidan
  Clark, AJ Ostrow, Akila Welihinda, Alan Hayes, Alec Radford, et~al.
\newblock Gpt-4o system card.
\newblock \emph{arXiv preprint arXiv:2410.21276}, 2024.

\bibitem[Jacob(2021)]{jacob2021pytorch}
Gildenblat Jacob.
\newblock Pytorch library for cam methods, 2021.

\bibitem[Jain et~al.(2021)Jain, Agrawal, Saporta, Truong, Duong, Bui, Chambon,
  Zhang, Lungren, Ng, Langlotz, and Rajpurkar]{Jain2021RadGraphEC}
Saahil Jain, Ashwin Agrawal, Adriel Saporta, Steven Truong, D. Duong, Tan Bui,
  Pierre Chambon, Yuhao Zhang, Matthew~P. Lungren, Andrew~Y. Ng, Curt~P.
  Langlotz, and Pranav Rajpurkar.
\newblock Radgraph: Extracting clinical entities and relations from radiology
  reports.
\newblock \emph{ArXiv}, abs/2106.14463, 2021.

\bibitem[Jin et~al.(2024)Jin, Che, Lin, and Chen]{jin2024promptmrg}
Haibo Jin, Haoxuan Che, Yi Lin, and Hao Chen.
\newblock Promptmrg: Diagnosis-driven prompts for medical report generation.
\newblock In \emph{Proceedings of the AAAI Conference on Artificial
  Intelligence}, pages 2607--2615, 2024.

\bibitem[Jing et~al.(2017)Jing, Xie, and Xing]{jing2017automatic}
Baoyu Jing, Pengtao Xie, and Eric Xing.
\newblock On the automatic generation of medical imaging reports.
\newblock \emph{arXiv preprint arXiv:1711.08195}, 2017.

\bibitem[Kipf(2016)]{kipf2016semi}
TN Kipf.
\newblock Semi-supervised classification with graph convolutional networks.
\newblock \emph{arXiv preprint arXiv:1609.02907}, 2016.

\bibitem[Li et~al.(2025)Li, Wang, Sun, He, and Feng]{li2025context}
Hongzhao Li, Hongyu Wang, Xia Sun, Hua He, and Jun Feng.
\newblock Context-enhanced framework for medical image report generation using
  multimodal contexts.
\newblock \emph{Knowledge-Based Systems}, 310:\penalty0 112913, 2025.

\bibitem[Li et~al.(2023{\natexlab{a}})Li, Li, Savarese, and Hoi]{li2023blip}
Junnan Li, Dongxu Li, Silvio Savarese, and Steven Hoi.
\newblock Blip-2: Bootstrapping language-image pre-training with frozen image
  encoders and large language models.
\newblock In \emph{International conference on machine learning}, pages
  19730--19742. PMLR, 2023{\natexlab{a}}.

\bibitem[Li et~al.(2023{\natexlab{b}})Li, Lin, Chen, Lin, Liang, and
  Chang]{li2023DCL}
Mingjie Li, Bingqian Lin, Zicong Chen, Haokun Lin, Xiaodan Liang, and Xiaojun
  Chang.
\newblock Dynamic graph enhanced contrastive learning for chest x-ray report
  generation.
\newblock In \emph{Proceedings of the IEEE/CVF Conference on Computer Vision
  and Pattern Recognition}, pages 3334--3343, 2023{\natexlab{b}}.

\bibitem[Li et~al.(2024)Li, Lin, Qiu, Liang, Chen, Elsaddik, and
  Chang]{li2024CoFE}
Mingjie Li, Haokun Lin, Liang Qiu, Xiaodan Liang, Ling Chen, Abdulmotaleb
  Elsaddik, and Xiaojun Chang.
\newblock Contrastive learning with counterfactual explanations for radiology
  report generation.
\newblock In \emph{European Conference on Computer Vision}, pages 162--180.
  Springer, 2024.

\bibitem[Liang et~al.(2024)Liang, Zhang, Wang, Zhong, Li, and
  Wang]{liang2024divide}
Xiao Liang, Yanlei Zhang, Di Wang, Haodi Zhong, Ronghan Li, and Quan Wang.
\newblock Divide and conquer: Isolating normal-abnormal attributes in knowledge
  graph-enhanced radiology report generation.
\newblock In \emph{ACM Multimedia 2024}, 2024.

\bibitem[Lin(2004)]{lin2004rouge}
Chin-Yew Lin.
\newblock Rouge: A package for automatic evaluation of summaries.
\newblock In \emph{Text summarization branches out}, pages 74--81, 2004.

\bibitem[Liu et~al.(2024{\natexlab{a}})Liu, Guo, Yong, and Xu]{liu2024multi}
Aohan Liu, Yuchen Guo, Jun-hai Yong, and Feng Xu.
\newblock Multi-grained radiology report generation with sentence-level
  image-language contrastive learning.
\newblock \emph{IEEE Transactions on Medical Imaging}, 43\penalty0
  (7):\penalty0 2657--2669, 2024{\natexlab{a}}.

\bibitem[Liu et~al.(2024{\natexlab{b}})Liu, Tian, Chen, Song, and
  Zhang]{liu2024boostrapping}
Chang Liu, Yuanhe Tian, Weidong Chen, Yan Song, and Yongdong Zhang.
\newblock Bootstrapping large language models for radiology report generation.
\newblock In \emph{Proceedings of the AAAI Conference on Artificial
  Intelligence}, pages 18635--18643, 2024{\natexlab{b}}.

\bibitem[Liu et~al.(2021{\natexlab{a}})Liu, Wu, Ge, Fan, and
  Zou]{liu2021exploring}
Fenglin Liu, Xian Wu, Shen Ge, Wei Fan, and Yuexian Zou.
\newblock Exploring and distilling posterior and prior knowledge for radiology
  report generation.
\newblock In \emph{Proceedings of the IEEE/CVF conference on computer vision
  and pattern recognition}, pages 13753--13762, 2021{\natexlab{a}}.

\bibitem[Liu et~al.(2021{\natexlab{b}})Liu, Yin, Wu, Ge, Zou, Zhang, and
  Sun]{liu2021contrastive}
Fenglin Liu, Changchang Yin, Xian Wu, Shen Ge, Yuexian Zou, Ping Zhang, and Xu
  Sun.
\newblock Contrastive attention for automatic chest x-ray report generation.
\newblock \emph{arXiv preprint arXiv:2106.06965}, 2021{\natexlab{b}}.

\bibitem[Liu et~al.(2021{\natexlab{c}})Liu, You, Wu, Ge, Sun,
  et~al.]{liu2021auto}
Fenglin Liu, Chenyu You, Xian Wu, Shen Ge, Xu Sun, et~al.
\newblock Auto-encoding knowledge graph for unsupervised medical report
  generation.
\newblock \emph{Advances in Neural Information Processing Systems},
  34:\penalty0 16266--16279, 2021{\natexlab{c}}.

\bibitem[Liu et~al.(2022)Liu, Ge, Zou, and Wu]{liu2022competence}
Fenglin Liu, Shen Ge, Yuexian Zou, and Xian Wu.
\newblock Competence-based multimodal curriculum learning for medical report
  generation.
\newblock \emph{arXiv preprint arXiv:2206.14579}, 2022.

\bibitem[Liu et~al.(2021{\natexlab{d}})Liu, Lin, Cao, Hu, Wei, Zhang, Lin, and
  Guo]{liu2021swin}
Ze Liu, Yutong Lin, Yue Cao, Han Hu, Yixuan Wei, Zheng Zhang, Stephen Lin, and
  Baining Guo.
\newblock Swin transformer: Hierarchical vision transformer using shifted
  windows.
\newblock In \emph{Proceedings of the IEEE/CVF international conference on
  computer vision}, pages 10012--10022, 2021{\natexlab{d}}.

\bibitem[Loshchilov and Hutter(2017)]{loshchilov2017decoupled}
Ilya Loshchilov and Frank Hutter.
\newblock Decoupled weight decay regularization.
\newblock \emph{arXiv preprint arXiv:1711.05101}, 2017.

\bibitem[Messina et~al.(2022)Messina, Pino, Parra, Soto, Besa, Uribe,
  And{\'\i}a, Tejos, Prieto, and Capurro]{messina2022survey}
Pablo Messina, Pablo Pino, Denis Parra, Alvaro Soto, Cecilia Besa, Sergio
  Uribe, Marcelo And{\'\i}a, Cristian Tejos, Claudia Prieto, and Daniel
  Capurro.
\newblock A survey on deep learning and explainability for automatic report
  generation from medical images.
\newblock \emph{ACM Computing Surveys (CSUR)}, 54\penalty0 (10s):\penalty0
  1--40, 2022.

\bibitem[Papineni et~al.(2002)Papineni, Roukos, Ward, and
  Zhu]{papineni2002bleu}
Kishore Papineni, Salim Roukos, Todd Ward, and Wei-Jing Zhu.
\newblock Bleu: a method for automatic evaluation of machine translation.
\newblock In \emph{Proceedings of the 40th annual meeting of the Association
  for Computational Linguistics}, pages 311--318, 2002.

\bibitem[Paszke et~al.(2019)Paszke, Gross, Massa, Lerer, Bradbury, Chanan,
  Killeen, Lin, Gimelshein, Antiga, et~al.]{paszke2019pytorch}
Adam Paszke, Sam Gross, Francisco Massa, Adam Lerer, James Bradbury, Gregory
  Chanan, Trevor Killeen, Zeming Lin, Natalia Gimelshein, Luca Antiga, et~al.
\newblock Pytorch: An imperative style, high-performance deep learning library.
\newblock \emph{Advances in neural information processing systems}, 32, 2019.

\bibitem[Rahman et~al.(2025)Rahman, Lee, Vu, Murtza, and Kim]{rahman2025duco}
Zahid~Ur Rahman, Ju-Hwan Lee, Dang~Thanh Vu, Iqbal Murtza, and Jin-Young Kim.
\newblock Duco-net: Dual-contrastive learning network for medical report
  retrieval leveraging enhanced encoders and augmentations.
\newblock \emph{IEEE Access}, 2025.

\bibitem[Schlichtkrull et~al.(2018)Schlichtkrull, Kipf, Bloem, Van Den~Berg,
  Titov, and Welling]{schlichtkrull2018RGCN}
Michael Schlichtkrull, Thomas~N Kipf, Peter Bloem, Rianne Van Den~Berg, Ivan
  Titov, and Max Welling.
\newblock Modeling relational data with graph convolutional networks.
\newblock In \emph{European semantic web conference}, pages 593--607. Springer,
  2018.

\bibitem[Touvron et~al.(2023{\natexlab{a}})Touvron, Martin, Stone, Albert,
  Almahairi, Babaei, Bashlykov, Batra, Bhargava, Bhosale,
  et~al.]{touvron2023llama}
Hugo Touvron, Louis Martin, Kevin Stone, Peter Albert, Amjad Almahairi, Yasmine
  Babaei, Nikolay Bashlykov, Soumya Batra, Prajjwal Bhargava, Shruti Bhosale,
  et~al.
\newblock Llama 2: Open foundation and fine-tuned chat models.
\newblock \emph{arXiv preprint arXiv:2307.09288}, 2023{\natexlab{a}}.

\bibitem[Touvron et~al.(2023{\natexlab{b}})Touvron, Martin, Stone, Albert,
  Almahairi, Babaei, Bashlykov, Batra, Bhargava, Bhosale,
  et~al.]{touvron2023llama2}
Hugo Touvron, Louis Martin, Kevin Stone, Peter Albert, Amjad Almahairi, Yasmine
  Babaei, Nikolay Bashlykov, Soumya Batra, Prajjwal Bhargava, Shruti Bhosale,
  et~al.
\newblock Llama 2: Open foundation and fine-tuned chat models.
\newblock \emph{arXiv preprint arXiv:2307.09288}, 2023{\natexlab{b}}.

\bibitem[Vedantam et~al.(2015)Vedantam, Lawrence~Zitnick, and
  Parikh]{vedantam2015cider}
Ramakrishna Vedantam, C Lawrence~Zitnick, and Devi Parikh.
\newblock Cider: Consensus-based image description evaluation.
\newblock In \emph{Proceedings of the IEEE conference on computer vision and
  pattern recognition}, pages 4566--4575, 2015.

\bibitem[Veli{\v{c}}kovi{\'c} et~al.(2017)Veli{\v{c}}kovi{\'c}, Cucurull,
  Casanova, Romero, Lio, and Bengio]{velivckovic2017graph}
Petar Veli{\v{c}}kovi{\'c}, Guillem Cucurull, Arantxa Casanova, Adriana Romero,
  Pietro Lio, and Yoshua Bengio.
\newblock Graph attention networks.
\newblock \emph{arXiv preprint arXiv:1710.10903}, 2017.

\bibitem[Wang et~al.(2022)Wang, Bhalerao, and He]{wang2022cross}
Jun Wang, Abhir Bhalerao, and Yulan He.
\newblock Cross-modal prototype driven network for radiology report generation.
\newblock In \emph{European Conference on Computer Vision}, pages 563--579.
  Springer, 2022.

\bibitem[Wang et~al.(2023{\natexlab{a}})Wang, Chen, Qian, Gao, Wei, Wang, Tian,
  and Gao]{wang2023MMPTMSurvey}
Xiao Wang, Guangyao Chen, Guangwu Qian, Pengcheng Gao, Xiao-Yong Wei, Yaowei
  Wang, Yonghong Tian, and Wen Gao.
\newblock Large-scale multi-modal pre-trained models: A comprehensive survey.
\newblock \emph{Machine Intelligence Research}, 20\penalty0 (4):\penalty0
  447--482, 2023{\natexlab{a}}.

\bibitem[Wang et~al.(2024{\natexlab{a}})Wang, Li, Wang, Wang, Li, and
  Jiang]{wang2024r2gencsr}
Xiao Wang, Yuehang Li, Fuling Wang, Shiao Wang, Chuanfu Li, and Bo Jiang.
\newblock R2gencsr: Retrieving context samples for large language model based
  x-ray medical report generation.
\newblock \emph{arXiv preprint arXiv:2408.09743}, 2024{\natexlab{a}}.

\bibitem[Wang et~al.(2024{\natexlab{b}})Wang, Li, Wu, Jin, Rong, Jiang, Li, and
  Tang]{wang2024pretrainXray}
Xiao Wang, Yuehang Li, Wentao Wu, Jiandong Jin, Yao Rong, Bo Jiang, Chuanfu Li,
  and Jin Tang.
\newblock Pre-training on high definition x-ray images: An experimental study.
\newblock \emph{arXiv preprint arXiv:2404.17926}, 2024{\natexlab{b}}.

\bibitem[Wang et~al.(2025{\natexlab{a}})Wang, Wang, Li, Ma, Wang, Jiang, and
  Tang]{Wang_2025_CVPR}
Xiao Wang, Fuling Wang, Yuehang Li, Qingchuan Ma, Shiao Wang, Bo Jiang, and Jin
  Tang.
\newblock Cxpmrg-bench: Pre-training and benchmarking for x-ray medical report
  generation on chexpert plus dataset.
\newblock In \emph{Proceedings of the Computer Vision and Pattern Recognition
  Conference (CVPR)}, pages 5123--5133, 2025{\natexlab{a}}.

\bibitem[Wang et~al.(2025{\natexlab{b}})Wang, Wang, Wang, Jiang, Li, Wang,
  Tian, and Tang]{wang2025AMMRG}
Xiao Wang, Fuling Wang, Haowen Wang, Bo Jiang, Chuanfu Li, Yaowei Wang,
  Yonghong Tian, and Jin Tang.
\newblock Activating associative disease-aware vision token memory for
  llm-based x-ray report generation.
\newblock \emph{arXiv preprint arXiv:2501.03458}, 2025{\natexlab{b}}.

\bibitem[Wang et~al.(2023{\natexlab{b}})Wang, Lin, and
  Dong]{Wang2023RethinkingMR}
Yixin Wang, Zihao Lin, and Haoyu Dong.
\newblock Rethinking medical report generation: Disease revealing enhancement
  with knowledge graph.
\newblock \emph{ArXiv}, abs/2307.12526, 2023{\natexlab{b}}.

\bibitem[Wang et~al.(2023{\natexlab{c}})Wang, Liu, Wang, and
  Zhou]{wang2023metransformer}
Zhanyu Wang, Lingqiao Liu, Lei Wang, and Luping Zhou.
\newblock Metransformer: Radiology report generation by transformer with
  multiple learnable expert tokens.
\newblock In \emph{Proceedings of the IEEE/CVF conference on computer vision
  and pattern recognition}, pages 11558--11567, 2023{\natexlab{c}}.

\bibitem[Wang et~al.(2023{\natexlab{d}})Wang, Liu, Wang, and
  Zhou]{wang2023r2gengpt}
Zhanyu Wang, Lingqiao Liu, Lei Wang, and Luping Zhou.
\newblock R2gengpt: Radiology report generation with frozen llms.
\newblock \emph{Meta-Radiology}, 1\penalty0 (3):\penalty0 100033,
  2023{\natexlab{d}}.

\bibitem[Xing et~al.(2025)Xing, Song, Zhang, Feng, Yu, and Yang]{xing2025mca}
Qilong Xing, Zikai Song, Youjia Zhang, Na Feng, Junqing Yu, and Wei Yang.
\newblock Mca-rg: Enhancing llms with medical concept alignment for radiology
  report generation.
\newblock \emph{arXiv preprint arXiv:2507.06992}, 2025.

\bibitem[Xue et~al.(2024)Xue, Tan, Tan, Qin, and Xiang]{xue2024generating}
Youyuan Xue, Yun Tan, Ling Tan, Jiaohua Qin, and Xuyu Xiang.
\newblock Generating radiology reports via auxiliary signal guidance and a
  memory-driven network.
\newblock \emph{Expert Systems with Applications}, 237:\penalty0 121260, 2024.

\bibitem[Yan et~al.(2021)Yan, He, Lu, Du, Chang, Gentili, McAuley, and
  Hsu]{yan2021weakly}
An Yan, Zexue He, Xing Lu, Jiang Du, Eric Chang, Amilcare Gentili, Julian
  McAuley, and Chun-Nan Hsu.
\newblock Weakly supervised contrastive learning for chest x-ray report
  generation.
\newblock \emph{arXiv preprint arXiv:2109.12242}, 2021.

\bibitem[Yan(2022)]{yan2022memory}
Sixing Yan.
\newblock Memory-aligned knowledge graph for clinically accurate radiology
  image report generation.
\newblock In \emph{Proceedings of the 21st Workshop on Biomedical Language
  Processing}, pages 116--122, 2022.

\bibitem[Yan et~al.(2023)Yan, Cheung, Chiu, Tong, Cheung, and
  See]{yan2023attributed}
Sixing Yan, William~K Cheung, Keith Chiu, Terence~M Tong, Ka~Chun Cheung, and
  Simon See.
\newblock Attributed abnormality graph embedding for clinically accurate x-ray
  report generation.
\newblock \emph{IEEE Transactions on Medical Imaging}, 42\penalty0
  (8):\penalty0 2211--2222, 2023.

\bibitem[Yang et~al.(2025)Yang, Li, Yang, Zhang, Hui, Zheng, Yu, Gao, Huang,
  Lv, et~al.]{yang2025qwen3}
An Yang, Anfeng Li, Baosong Yang, Beichen Zhang, Binyuan Hui, Bo Zheng, Bowen
  Yu, Chang Gao, Chengen Huang, Chenxu Lv, et~al.
\newblock Qwen3 technical report.
\newblock \emph{arXiv preprint arXiv:2505.09388}, 2025.

\bibitem[Yang et~al.(2024)Yang, Yu, Fu, Zhang, Yu, Wang, Jiang, Lv, Huang, and
  Han]{yang2024token}
Yan Yang, Jun Yu, Zhenqi Fu, Ke Zhang, Ting Yu, Xianyun Wang, Hanliang Jiang,
  Junhui Lv, Qingming Huang, and Weidong Han.
\newblock Token-mixer: Bind image and text in one embedding space for medical
  image reporting.
\newblock \emph{IEEE Transactions on Medical Imaging}, 43\penalty0
  (11):\penalty0 4017--4028, 2024.

\bibitem[You et~al.(2021)You, Liu, Ge, Xie, Zhang, and
  Wu]{you2021aligntransformer}
Di You, Fenglin Liu, Shen Ge, Xiaoxia Xie, Jing Zhang, and Xian Wu.
\newblock Aligntransformer: Hierarchical alignment of visual regions and
  disease tags for medical report generation.
\newblock In \emph{International Conference on Medical Image Computing and
  Computer-Assisted Intervention}, pages 72--82. Springer, 2021.

\bibitem[Zhang et~al.(2024)Zhang, Acosta, Zhou, and
  Rajpurkar]{zhang2024uncovering}
Xiaoman Zhang, Juli{\'a}n~N Acosta, Hong-Yu Zhou, and Pranav Rajpurkar.
\newblock Uncovering knowledge gaps in radiology report generation models
  through knowledge graphs.
\newblock \emph{arXiv preprint arXiv:2408.14397}, 2024.

\bibitem[Zhang et~al.(2020)Zhang, Wang, Xu, Yu, Yuille, and
  Xu]{zhang2020radiology}
Yixiao Zhang, Xiaosong Wang, Ziyue Xu, Qihang Yu, Alan Yuille, and Daguang Xu.
\newblock When radiology report generation meets knowledge graph.
\newblock In \emph{Proceedings of the AAAI conference on artificial
  intelligence}, pages 12910--12917, 2020.

\end{thebibliography}
}


\end{document}